\documentclass[10pt,twocolumn,letterpaper]{article}

\usepackage{cvpr}              

\usepackage{graphicx}
\usepackage{amsmath}
\usepackage{amssymb}
\usepackage{booktabs}

\usepackage[accsupp]{axessibility}  

\usepackage[pagebackref,breaklinks,colorlinks]{hyperref}

\usepackage[capitalize]{cleveref}
\crefname{section}{Sec.}{Secs.}
\Crefname{section}{Section}{Sections}
\Crefname{table}{Table}{Tables}
\crefname{table}{Tab.}{Tabs.}

\usepackage{booktabs}
\usepackage{multirow}

\def\cvprPaperID{10781} 
\def\confName{CVPR}
\def\confYear{2023}

\begin{document}

\title{DSFNet: Dual Space Fusion Network for Occlusion-Robust \\
3D Dense Face Alignment}

\author{Heyuan Li$^1$, Bo Wang$^2$, Yu Cheng$^1$, Mohan Kankanhalli$^1$, Robby T. Tan$^{1}$\\
$^1$ National University of Singapore \hspace{1mm} $^2$CtrsVision \hspace{1mm}\\
\tt\small liheyuan@u.nus.edu, hawk.rsrch@gmail.com, e0321276@u.nus.edu,\\
\tt\small mohan@comp.nus.edu.sg, robby.tan@nus.edu.sg
}

\twocolumn[{%
\renewcommand\twocolumn[1][]{#1}%
\maketitle

\begin{center}
\centering
    {\includegraphics[width=1.0\linewidth]{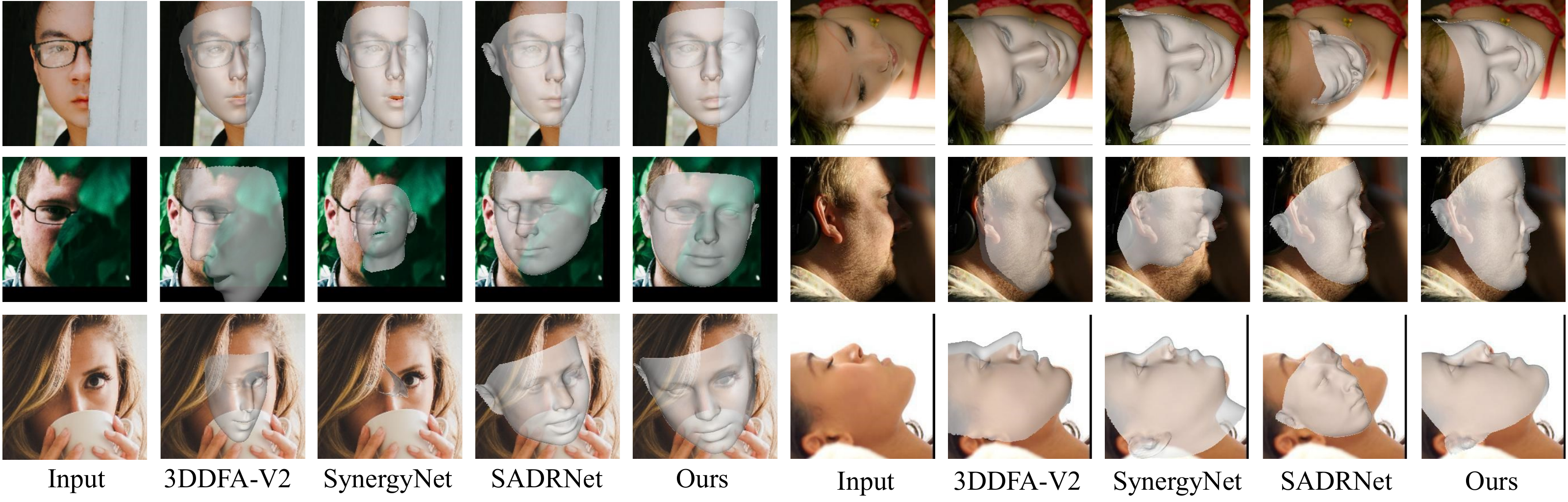}}
    \captionof{figure}{The qualitative results compared to state-of-the-art methods on severely occluded or large view angle cases from datasets \cite{zhu2016face,voo2022delving}.  
    For each case, from left to right are input, results of 3DDFA-V2 \cite{guo2020towards}, SynergyNet \cite{wu2021synergy}, SADRNet \cite{ruan2021sadrnet}, and our method.}
    \label{fig:vis}
\end{center}

}]

\begin{abstract}
Sensitivity to severe occlusion and large view angles limits the usage scenarios of the existing monocular 3D dense face alignment methods.
The state-of-the-art 3DMM-based method, directly regresses the model's coefficients, underutilizing the low-level 2D spatial and semantic information, which can actually offer cues for face shape and orientation.
In this work, we demonstrate how modeling 3D facial geometry in image and model space jointly can solve the occlusion and view angle problems. 
Instead of predicting the whole face directly, we regress image space features in the visible facial region by dense prediction first. 
Subsequently, we predict our model's coefficients based on the regressed feature of the visible regions, leveraging the prior knowledge of whole face geometry from the morphable models to complete the invisible regions. 
We further propose a fusion network that combines the advantages of both the image and model space predictions to achieve high robustness and accuracy in unconstrained scenarios. 
Thanks to the proposed fusion module, our method is robust not only to occlusion and large pitch and roll view angles, which is the benefit of our image space approach, but also to noise and large yaw angles, which is the benefit of our model space method.
Comprehensive evaluations demonstrate the superior performance of our method compared with the state-of-the-art methods.
On the 3D dense face alignment task, we achieve \textbf{3.80\%} NME on the AFLW2000-3D dataset, which outperforms the state-of-the-art method by \textbf{5.5\%}. 
Code is available at \url{https://github.com/lhyfst/DSFNet}.
\end{abstract}

\section{Introduction}
\label{sec:intro}

3D dense face alignment is an important problem with many applications, e.g. video conferencing, AR/VR/metaverse, games, facial analysis, etc. 
Many methods have been proposed~\cite{tran2017regressing, feng2018prn, deng2019accurate, DECA:Siggraph2021, shang2020self, zhu2016face, jourabloo2016large, wu2021synergy, ruan2021sadrnet, guo2020towards, zielonka2022towards, wen2021self, dib2022s2f2, sanyal2019learning, meng20223d}. However, these methods are sensitive to severe occlusion and large view angles \cite{li2021fit, tiwari2022occlusion, tiwari2021self, ruan2021sadrnet}, limiting their  applicability of 3D dense face alignment on wild images where occlusion and view angles often occur.

3D dense face alignment from a single image is an ill-posed problem, mainly because of the depth ambiguity. 
The existing methods~\cite{tran2017regressing, deng2019accurate, shang2020self, DECA:Siggraph2021, sanyal2019learning} use a contractive CNN to predict the coefficients of 3DMM~\cite{blanz1999morphable} directly. However, contractive CNNs are essentially ill-suited for this task~\cite{koizumi2020look} due to some reasons, including: mixed depth and  2D spatial information, the loss of low-level 2D spatial information as a result of invariance attribute of CNNs, and mixed facial region and occluder in the process of the contraction.

\begin{figure}[t]
  \centering
   \includegraphics[width=1.0\linewidth]{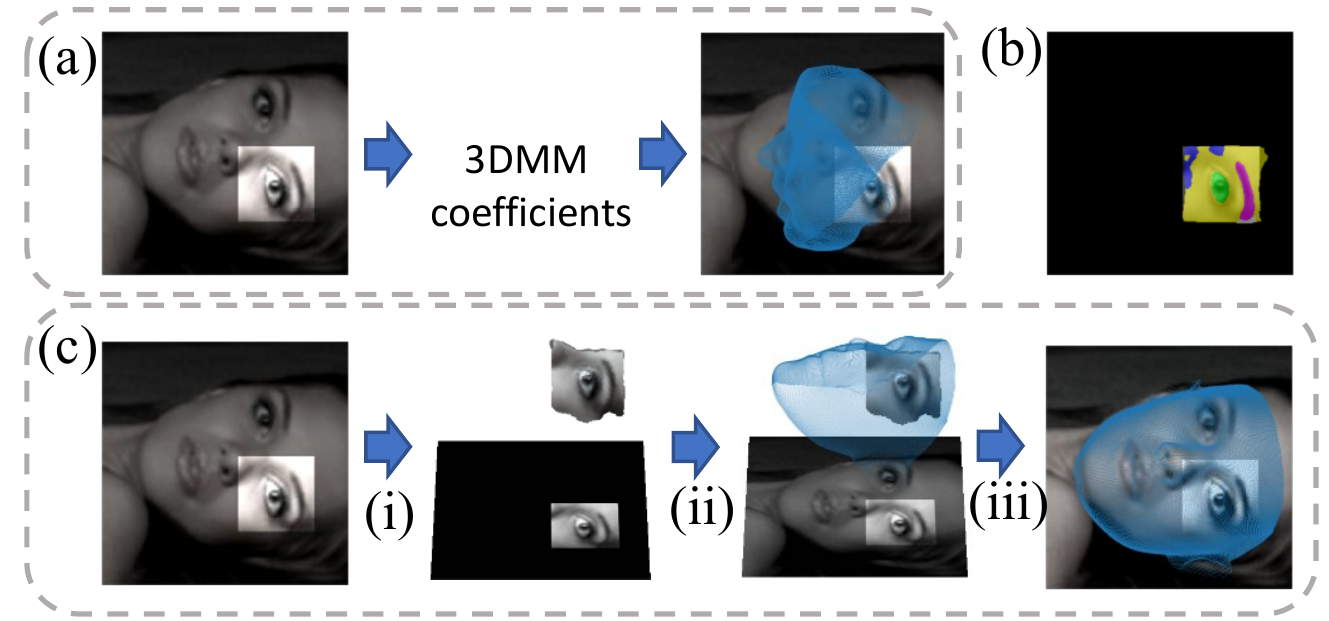}
   \caption{In this case, only one eye is visible. (a) The 3DMM-based method fails. (b) Face parsing algorithm \cite{yu2018bisenet} still works. (c) Our method first (i) predicts reliable geometry in visible region by dense prediction, then (ii) completes the whole face by facial geometry prior, producing a reasonable result. (iii) Viewed in image view.}
   \label{fig:intro}
\end{figure}

Severe occlusion and large view angles pose problems due to the complexity of the many-to-one mapping from 2D image to 3D shape. In contrast, low-level vision features are less variant according to geometry transform. Therefore, dense prediction is essentially more robust to the above problem in the visible region, because dense prediction relies more on local information, where an example is shown in \cref{fig:intro} (b). Even if most of the face is masked out and only the left eye is visible, the face parsing algorithm is still able to deduce a reasonable parsing result.

Based on this observation, we decentralize the instance-level 3DMM coefficients regression (i.e., whole-face level) to pixel-level dense prediction in image space to improve the robustness against occlusion and large view angles, by proposing a 3D facial geometry's 2D image space representation. 
To complete the invisible region due to extra- or self-occlusion, a novel post-process algorithm is proposed to convert the dense prediction for the visible face region into 3D facial geometry that includes the whole face area. 
\cref{fig:intro} (c) shows that our image space prediction recovers reasonable results only seeing one eye, while the SOTA method fails to produce a reasonable result.

We further compare the robustness and accuracy between the image space prediction with the model space prediction that directly regresses 3DMM's coefficients, and discover that there is a complementary relationship between these two spaces. 
Thus, we propose a dual space fusion network (DSFNet) that predicts using the image and model spaces using a two-branch architecture. With the fusion module, our DSFNet effectively combines the advantages of both spaces. In summary, the main contributions of this paper are:

\begin{itemize}
    \item We propose a novel 3D facial geometry's 2D image space representation, followed by a novel post-processing  algorithm. It achieves robust 3D dense face alignment to occlusion and large view angles. 
    \item We introduce a fusion network, which combines the advantages of both the image and model space predictions to achieve high robustness and accuracy in unconstrained scenarios.
    \item On the 3D dense face alignment task, we achieve \textbf{3.80\%} NME on AFLW2000-3D dataset, which outperforms the state-of-the-art method by \textbf{5.5\%}. 
\end{itemize}

\section{Related Works}
\label{sec:rela}

\begin{figure*}
\centering
    {\includegraphics[width=1.0\linewidth]{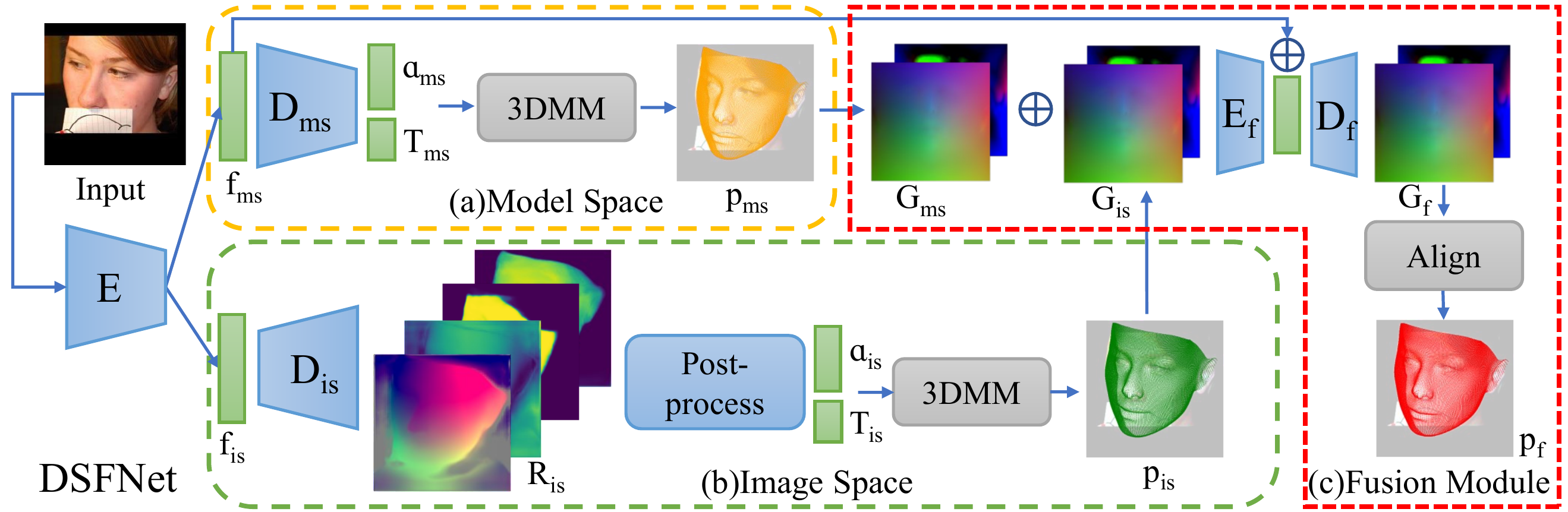}}
    \caption{The framework of our proposed Dual Space Fusion Network (DSFNet).}
    \label{fig:main_pipeline}
\end{figure*}

\noindent \textbf{3D Dense Face Alignment} 
3D facial geometry is a finite deformable object, which has specific distributions according to gender, age, and ethnicity. Since 3DMM compactly represents such prior knowledge by Principal Component Analysis (PCA), 3DMM model-based methods have been dominating 3D facial geometry modeling since deep learning revived in the past decade. A batch  of works \cite{feng2018prn, wu2021synergy, zhu2016face, guo2020towards, zielonka2022towards} directly supervise the deep learning model by ground truth 3DMM coefficients in annotated datasets. Another batch  of works \cite{deng2019accurate, DECA:Siggraph2021, shang2020self, wen2021self, dib2022s2f2, sanyal2019learning} supervise the deep learning model by photometric information in unlabelled images by taking advantage of differentiable rendering in a self-supervised way. 

On the contrary, there are some model-free methods \cite{feng2018prn, ruan2021sadrnet, jackson2017large, zhu2020beyond} that do not involve 3DMM, but directly predict 3D coordinates of vertices of the face mesh. Therefore, they theoretically have more geometry representation flexibility. PRN \cite{feng2018prn} and SADRNet \cite{ruan2021sadrnet} predict a position map in UV space. VRN \cite{jackson2017large} outputs its prediction in 3D volumetric space, which leads to large parameter volume and heavy computing burden. Model-free methods always are able to give precise predictions in visible region, but cannot tackle invisible region well, because of lack of the prior knowledge of 3D facial geometry distribution.

\vspace{0.2cm}
\noindent \textbf{Occlusion-Aware 3D Dense Face Alignment} 
Recently, more and more works \cite{tran2017extreme, ruan2021sadrnet, li2021fit} realize the importance of robustness to occlusion and large view angles which are common in unconstrained images. Some methods enhanced their robustness by focusing on the visible facial region. SADRNet \cite{ruan2021sadrnet} proposes an attention mechanism that lets the model only focus on the visible facial region. \cite{li2021fit} proposed a precise face skin segmentation network which masks out non-facial regions, so that it supervises only the model-based encoder by photometric signal in the visible face skin region. Another series of works achieve higher robustness to occlusion by data augmentation. \cite{tiwari2021self} proposed a Self-Supervised Robustifying Guidance framework which takes advantage of the consistency between randomly occluded images and their corresponding unoccluded images. \cite{tiwari2022occlusion} further proposed a Multi-Occlusion Per Identity framework which takes advantage of the consistency between a batch of images from the same identity but occluded with different random occlusion patterns.

Another batch of works \cite{alp2017densereg, yu2017learning, crispell2017pix2face, zhu2016face, koizumi2020look, sela2017unrestricted, wood20223d, kao2022single} predict visible facial geometry in image space first, and the invisible region can be recovered from the predicted visible region using prior knowledge in 3DMM, which is always conducted by an optimization-based algorithm. DenseReg \cite{alp2017densereg} proposed a correspondence map that contains the UV coordinates of vertices on the face model rasterized to the image plane. \cite{yu2017learning} predicts the correspondences between the input image and a rendered image of the 3DMM mean face geometry in the positive view by predicting 2D flow between them. \cite{crispell2017pix2face} represents the 3D facial geometry by a projected normalized coordinated code (PNCC) \cite{zhu2016face} and a 3D offset image. \cite{koizumi2020look} realizes that image-to-image CNNs are more suitable to predict the correspondence map because they reserve more low-level spatial information. \cite{sela2017unrestricted} represents the face by a depth map and a correspondence map, where each pixel represents its x, y, and z coordinates of the corresponding point on a normalized canonical face.  Some recent works don't use correspondence maps. 

Wood et al.~\cite{wood20223d} use hundreds of dense 2D landmarks as the image space representation. \cite{kao2022single} predicts the image space 2D-3D correspondence which is used to align predicted shape from canonical view to image view by Perspective-n-Point algorithm. However, these works have two limitations. First, some works have inferior accuracy, which is caused by poor 2D representations. Second, most works use optimization-based algorithms to post-process the 2D representation, but optimization-based algorithms bring more hyper-parameters which are tricky to tune and slow down the whole pipeline, because optimization-based algorithms update parameters iteratively. By contrast, our method uses a well-designed 3D facial geometry's 2D image space representation to reserve enough geometry information for accuracy and uses a lightweight PointNet-based module as the post-process algorithm instead of an optimization method, therefore avoiding burdened computing.

\section{Proposed Method}
\label{sec:meth}
We present DSFNet, our dual space fusion network, which aims at robust and accurate 3D facial geometry modeling by taking advantages from both model space prediction and image space prediction. 
Fig.~\ref{fig:main_pipeline} shows our DSFNet's architecture.
Our method takes a single face image as input, and uses HRNet \cite{SunXLW19} as an encoder $E$ to extract model space and image space features. 
The two space features are processed by model space branch and image space branch separately. Finally, the predictions from two branches are fed into a fusion module for the final prediction. 

\vspace{-0.3cm}
\paragraph{Facial Geometry Prior}
\label{subsec:meth_1}
We adopt the 3D Morphable Model as the geometry prior to our 3D face modeling. 
3DMM represents face shape $S \in \mathbb{R}_{3n}$ with $n$ vertices by identity and expression PCA bases and coefficients:
\begin{equation}
S = \bar{S} + B_{\rm id}\alpha_{\rm id} + B_{\rm exp}\alpha_{\rm exp} ,
\end{equation}
where $\bar{S}$ is the mean face geometry; $B_{\rm id}$ and $B_{\rm exp}$ are the PCA bases of identities and expressions. Variables $\alpha_{\rm id}$ and $\alpha_{\rm exp}$ are the corresponding coefficient vectors. We adopt the widely-used 2009 Basel Face Model \cite{paysan20093d}, where $\alpha_{\rm shape} \in \mathbb{R}_{199}$ and $\alpha_{\rm exp} \in \mathbb{R}_{29}$. 
Orthographic projection is used to align the 3D face shape to the image view:
$
G = fRS + t,
$
where $G \in \mathbb{R}_{3n}$ is aligning with the image view. Variables $f$, $R$ and $t$ are scale factor, 3D rotation matrix and 3D translation, respectively. We combine these three transformation parameters together as a pose vector $T \in \mathbb{R}_{12}$.

\subsection{Model Space Prediction}
\label{subsec:meth_2}
As shown in \cref{fig:main_pipeline} (a), we use a contractive CNN module as a model space decoder $D_{\rm ms}$ to regress model space prediction. 
The network takes $f_{\rm ms}$ as input, which is the smallest feature map ($8 \times 8$) from last stage of encoder $E$, and output the 3DMM coefficients $\alpha_{\rm ms}$ and pose vector $T_{\rm ms}$:

\begin{equation}
\alpha_{\rm ms}, T_{\rm ms} = D_{\rm ms}(f_{\rm ms}).
\end{equation}
With ground truth notation $\ast$ hereafter, the model space loss is shown as follows:
\begin{equation}
L_{\rm ms} = w_\alpha \left \| \alpha_{\rm ms} - \alpha^{\ast} \right \| ^{2} + w_{T}\left \| T_{\rm ms} - T^{\ast} \right \| ^{2},
\end{equation}
where $w$ denotes the weight of the corresponding item.

\begin{figure}[t]
  \centering
   \includegraphics[width=0.95\linewidth]{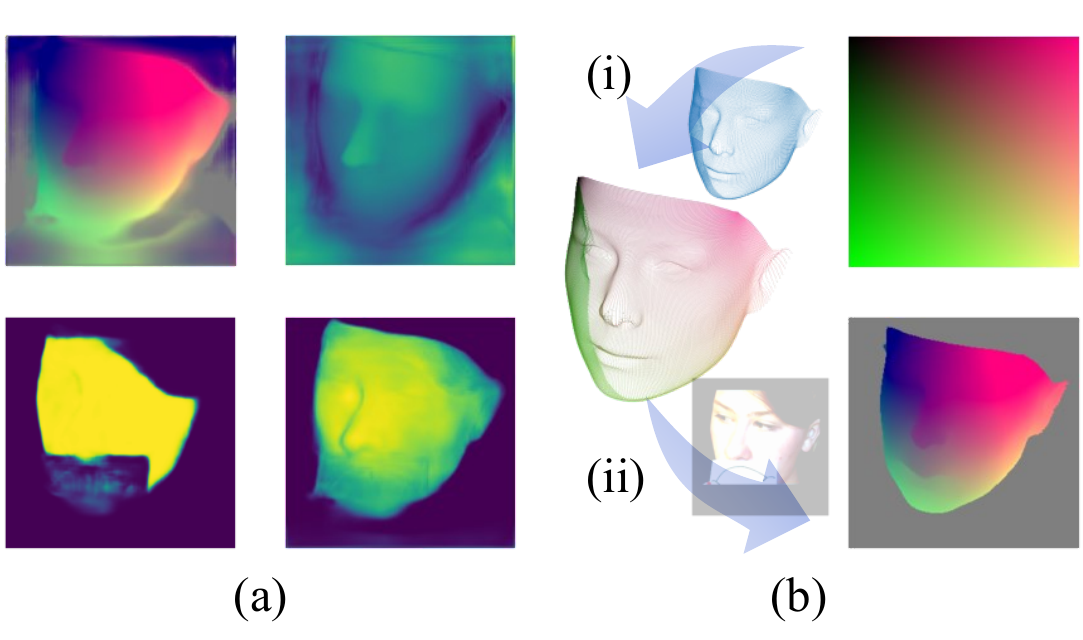}

   \caption{(a) An example of the four components of the image space representation in the inference stage. From left to right, from top to down, the four maps are correspondence map, depth map, segmentation map, and confidence map. (b) The illustration of the construction of a correspondence map. (i) Register UV locations to their corresponding 3D points. (ii) Rasterize the UV locations to the image plane to obtain the correspondence map.}
   \label{fig:image_space_representation}
\end{figure}

\subsection{Image Space Prediction}
\label{subsec:meth_3}
As depicted in \cref{fig:main_pipeline} (b), an image space decoder $D_{\rm is}$ takes $f_{\rm is}$ as input, which is the feature map ($64 \times 64$) concatenated from 4 scales of the output from the last stage of encoder $E$.
$D_{\rm is}$ predicts the image space representation $R_{\rm is}$ in the input size via upsampling and convolutions layers.
Then, a PointNet-based post-process algorithm $\rm Post$ converts the predicted representation into 3DMM coefficients $\alpha_{\rm is}$ and pose vector ${T_{\rm is}}$: 
\begin{eqnarray}
R_{\rm is} &=& D_{\rm is}(f_{\rm is}),\\
\alpha_{\rm is}, T_{\rm is} &=& {\rm Post}(R_{\rm is}).
\end{eqnarray}

\subsubsection{Image Space Representation}
Our image space representation $R_{\rm is}$ consists of four 2D maps: correspondence map $\rm Cor$, depth map $\rm Dep$, segmentation map $\rm Seg$, and confidence map $\rm Cf$, shown in \cref{fig:image_space_representation} (a). 
We represent a pixel $X$ in the input image by $(x,y) \in [1,H] \times [1,W] $, where $H$ and $W$ are the height and width of the input image $I$.

\vspace{-0.3cm}
\paragraph{Correspondence Map} 
$\rm Cor \in \mathbb{R}_{2 \times H \times W}$ represents pixels' UV coordinates on the 3DMM. 
$\rm Cor(x,y)$ bridges the correspondence between $X$'s location $(x,y)$ in image and its UV location $(u,v)$ on our face model. 
As shown in \cref{fig:image_space_representation} (b), our correspondence map is constructed in the following steps.
First, similar to representing a mesh's texture by a 2D texture map, we unwrap the 3DMM to a 2D UV space $M \in \mathbb{R}_{2 \times h \times w }$, where $h$ and $w$ are the height and width of the 2D map. 
Now, every vertex has a UV location $(u,v) \in [1,h] \times [1,w]$. 
The UV location is then normalized to $(u,v) \in [-1,1] \times [-1,1]$.
Every vertex's UV location is registered to its 3D location on $G$, and the correspondence map is obtained by rasterizering the UV locations to the image plane via $G$. 

\vspace{-0.3cm}
\paragraph{Depth Map} 
${\rm Dep} \in \mathbb{R}_{H \times W}$ represents pixels' relative depth, which contains the geometry information in the non-occluded region. Using relative depth is feasible due to orthogonal projection, which helps the model by narrowing the prediction space. The nose tip's depth is used as zero point to compute the relative depth.

\vspace{-0.3cm}
\paragraph{Segmentation Map} 
${\rm Seg} \in \mathbb{R}_{H \times W}$ indicates the visible facial regions. 
Its values range from 0 to 1. 
The larger the value is, the more certain the pixel in visible facial regions is. 
The visible region $V$ is defined as all pixels where $\rm Seg(x,y)>\theta$. 
$\theta $ denotes a threshold set to 0.5 in our experiments.

\vspace{-0.3cm}
\paragraph{Confidence Map} 
${\rm Cf} \in \mathbb{R}_{H \times W}$ represents the reliability of $\rm Cor$ and $\rm Dep$. Since there is no ground truth of it,  we compute the ground truth $\rm Cf(x,y)^{\ast }$ as follows:
\begin{equation}
\label{eq:con_map}
{\rm Cf}(x,y)^{\ast} = \sqrt{{\rm Cf}_{\rm Cor}(x,y)^{\ast }{\rm Cf}_{\rm Dep}(x,y)^{\ast }}, 
\end{equation}
where:
\begin{small}
\begin{eqnarray}
{\rm Cf}_{\rm Cor}(x,y)^{\ast } &=& \exp\left(-\frac{\left | {\rm Cor}(x,y) - {\rm Cor}(x,y)^{\ast }  \right | }{a} \right),\\
{\rm Cf}_{\rm Dep}(x,y)^{\ast } &=& \exp\left(-\frac{\left | {\rm Dep}(x,y) - {\rm Dep}(x,y)^{\ast }  \right | }{b} \right),
\end{eqnarray}
\end{small} 
\hspace{-0.18cm} with ${\rm Cf}_{\rm Cor}(x,y)^{\ast }$ and ${\rm Cf}_{\rm Dep}(x,y)^{\ast }$ denote the ground truth confidence values of ${\rm Cor}(x,y)$ and ${\rm Dep}(x,y)$, respectively. Variables $a$ and $b$ are parameters that control the tolerance to the error of predicted ${\rm Cor}(x,y)$ and ${\rm Dep}(x,y)$, respectively. 
\cref{fig:image_space_representation} illustrates that the confidence map effectively learns that facial attributes region is more reliable, but UV-island boundaries \cite{wood20223d} and side face area are less reliable.

The image space representation loss $L_{\rm Rep}$ is computed as follows:
\begin{equation}
L_{\rm Rep} = w_1 L_{\rm Cor} + w_2 L_{\rm Dep} + w_3 L_{\rm Seg} + w_4 L_{\rm Cf},
\end{equation}
where:
\begin{small}
\begin{eqnarray}
L_{\rm Cor} \hspace{-0.3cm}&=& \hspace{-0.3cm}\frac{1}{s(V)} \sum_{(x,y) \in V}\left \| {\rm Cor}(x,y) - {\rm Cor}(x,y)^{\ast} \right \| _{2},\\
L_{\rm Dep}\hspace{-0.3cm} &=& \hspace{-0.3cm}\frac{1}{s(V)}  \sum_{(x,y) \in V}\left \| {\rm Dep}(x,y) - {\rm Dep}(x,y)^{\ast} \right \| _{2},\\
L_{\rm Seg}\hspace{-0.3cm} &=&\hspace{-0.3cm} \frac{1}{HW}  \sum_{(x,y) \in I} \left \| {\rm Seg}(x,y) - {\rm Seg}(x,y)^{\ast} \right \|  _{2},\\
L_{\rm Cf}\hspace{-0.3cm} &=&\hspace{-0.3cm} \frac{1}{s(V)}  \sum_{(x,y) \in V}\left \| {\rm Cf}(x,y) - {\rm Cf}(x,y)^{\ast} \right \| _{2},
\end{eqnarray}
\end{small}
\hspace{-0.2cm} with $s(V)$ denotes the area of $V$.
$L_{\rm Seg}$ teaches the network to extract facial region from non-facial region. $L_{\rm Dep}$ teaches the network to learn and represent visible facial geometry. $L_{\rm Cor}$ teaches the network how to register visible facial geometry to the face model. With $L_{\rm Cf}$, the network learns the reliability of the predicted ${\rm Cor}$ and ${\rm Dep}$.

\begin{figure}[t]
  \centering
   \includegraphics[width=1\linewidth]{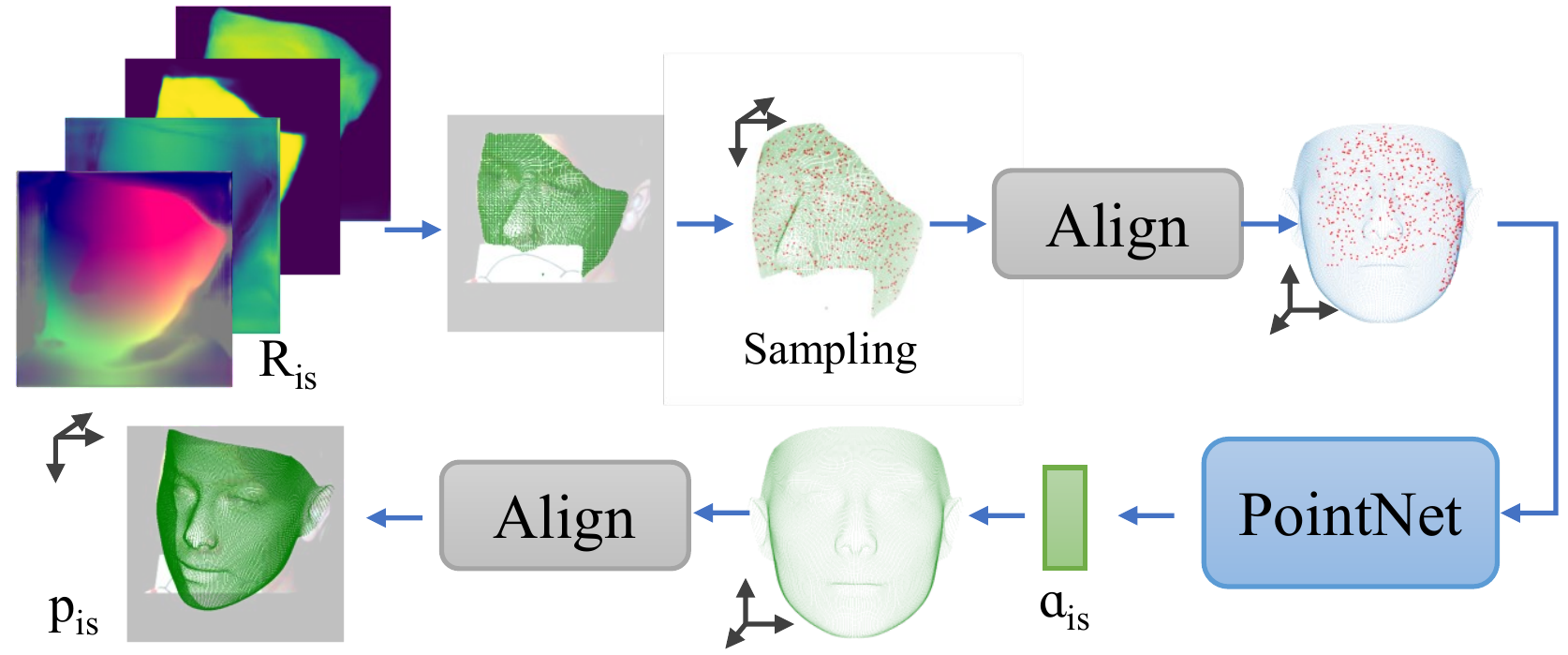}

   \caption{The post-process algorithm}
   \label{fig:post_process}
\end{figure}

\subsubsection{A PointNet-based Post-process Algorithm}
The image space representation contains only the predicted geometry in the visible regions. To infer the geometry in the invisible regions due to extra-occlusion or self-occlusion, we need to leverage the prior knowledge of 3D facial geometry in 3DMM. 

Previous works use optimization-based post-process algorithms to achieve the conversion. However, the optimization-based algorithms are slower due to their iterative updating parameters. 
Moreover, the optimized results might have distortion or might be too rigid due to the inappropriate weight of the regularization items.
To avoid these problems, we convert the image space representation into a point cloud format.
We then propose a PointNet-based post-process algorithm, which is fast and reliable. Illustrated in \cref{fig:post_process}, the algorithm has four steps: 

\vspace{-0.3cm}
\paragraph{Step 1: Convert Image Space Representation to Point Cloud} 
We randomly sample $m$ pixels from the visible regions $V$. 
Using orthographic projection, the 3D location $P$ of every selected pixel $X$ in the image view can be represented as $(x,y,z)$, where $z$ is ${\rm Dep}(x,y)$. $P_c$ denotes the point cloud consisting of these selected points.

\paragraph{Step 2: Align to Canonical View} 
\cite{ruan2021sadrnet} proposes a self-alignment module to align the face mesh to the image view by 68 facial landmarks. We use a similar algorithm $\rm Align$ to align the point cloud from the image view to the canonical view, but using a set of randomly selected points:
\begin{equation}
\hat{K}, T = {\rm Align}(K,D,W),
\end{equation}
where $K$ and $\hat{K}$ are the point clouds before and after the alignment, $T$ is the transform vector of the alignment, $D$ is the goal point cloud used to align to, $W$ is the corresponding weights.
We can find the values of $P$ corresponding to the  3D point $\bar{P}$ on a mean face's geometry in the canonical view by searching the point that has the nearest UV location to ${\rm Cor}(x,y)$. 
The corresponding point cloud on the mean face is denoted as $\bar{P}_c$. At the same time, ${\rm Cf}(x,y)$ indicates the reliability of $P$. 
Hence, the weight set of $P_c$ is denoted as ${\rm Cf}(P_c)$. Therefore, the transformed point cloud in the canonical view $\hat{P}_c$ and the transformed vector from the image view to the canonical view $T_c$ are represented as:
\begin{equation}
\hat{P}_c, T_c = {\rm Align}(P_c,\bar{P}_c, {\rm Cf}(P_c)).
\end{equation}

\paragraph{Step 3: Compute  3DMM Coefficients from Point Cloud} 
PointNet~\cite{qi2017pointnet} is a commonly used model to process point clouds. 
We here propose a PointNet-based module that converts the point cloud into the 3DMM geometry coefficients $\alpha_{is}$. 
Since the input point cloud $\hat{Pc}$ is in the canonical view, the transformation modules in the PointNet are removed, which makes the model more lightweight. 
For every point $\hat{P}$, we concatenate its 3D location with ${\rm Cor}(x,y)$ and ${\rm Cf}(x,y)$ as the input of the PointNet-based module $\rm Pnet$. This process is represented as:
\begin{equation}
\alpha_{\rm is} = {\rm Pnet}(\hat{P}_c, {\rm Cor}(P_c), {\rm Cf}(P_c)) .
\end{equation}

A previous work \cite{wu2021synergy} also makes use of the  PointNet-based module to regress the 3DMM coefficients and transform vectors by 3D points. 
Their module however only makes use of 68 sparse facial landmarks. 
Unlike this method, our module makes use of all the point clouds on the facial geometry in the visible regions, which contains much more geometry information. 
Our method separates the prediction of $T_{\rm is}$ and $\alpha_{\rm is}$ by the previous alignment algorithm and this PointNet-based module instead of mixing them together. Our ablation studies in \cref{sec:abla_post} demonstrates this disentanglement helps both of them.

\paragraph{Step 4: Align to Image View} As $\alpha_{\rm is}$ is obtained, the whole face mesh in the canonical view $\hat{G_{\rm is}}$ can be generated by 3DMM. The 3D facial geometry in the image view $G_{\rm is}$ can be acquired by aligning $\hat{G_{\rm is}}$ back to the image view:
\begin{eqnarray}
P_c^{\prime}, T_i &=& {\rm Align}(\hat{P}_c,P_c, {\rm Con}(P_c)),\\
G_{\rm is} &=& T_{i}^{m} \hat{G_{\rm is}},
\end{eqnarray}
where $P_c^{\prime}$ is the point cloud which is transformed back to image view, $T_{i}^{m}$ is the transform matrix reshaped from $T_{i}$ which is the transformed vector from canonical view to image view. 

The loss of training the PointNet-based post-process module is then expressed as:
\begin{eqnarray}
L_{\rm post} &=& w_{\alpha} \left \| \alpha_{\rm is} - \alpha^{\ast} \right \| ^{2} + w_5 L_{\rm cons},\\
L_{\rm cons} &=& \frac{1}{s(V)} \sum_{P \in V}^{} \left \| P - P^{\prime} \right \|^{2},
\end{eqnarray}
where $L_{\rm cons}$ is the geometry consistency in the visible facial region between all 3D points in $V$ and their corresponding 3D points $P^{\prime}$ in the image space prediction.
In the training phase, the image space loss consists of two items, image space representation loss $L_{\rm Rep}$ and post-process loss $L_{\rm Post}$ as: $L_{\rm is} = L_{\rm Rep} + L_{\rm Post}$.

\subsection{Dual Space Fusion}
\label{subsec:meth_4}
In our investigation, we realize that model space prediction and image space prediction are complementary (\cref{sec:dual_space}). Hence, it is possible to fuse them together by a fusion module in order to take advantage of both of them. 
We fuse the predictions from the two branches in the UV space in a model-free way. 
%
Similar to \cite{ruan2021sadrnet}, we represent a 3D face geometry $G$ by an offset map $\rm Off$ which indicates the offset of the face's shape from the mean face shape and a position map $\rm Pos$,  which indicates the face's pose in image view:
\begin{eqnarray}
S &=& \bar{S} + \rm Off, \\
G &=& {\rm Align}(S, \rm Pos,Cf).
\end{eqnarray}

The fusion module, as shown in \cref{fig:main_pipeline} (c) ($E_{f}$ and $D_{f}$), is a lightweight U-Net \cite{ronneberger2015u}, which has two heads for predicting fused offset map ${\rm Off}_{f}$ and fused position map ${\rm Pos}_{f}$ separately. Model space feature $f_{\rm ms}$ is concatenated with the feature map of the center layer of the U-Net, in order to be aware of which branch's prediction is more reliable. The fused prediction $G_{f}$ is represented as:
\begin{small}
\begin{eqnarray}
{\rm Off}_{f}, {\rm Pos}_{f} &=& F({\rm Off}_{\rm ms},{\rm Pos}_{\rm ms},{\rm Off}_{\rm is},{\rm Pos}_{\rm is},f_{\rm ms}), \\
G_{f} &=& {\rm Align}(\bar{S}+{\rm Off}_{f},{\rm {Pos}_{f},{\rm Cf})}.
\end{eqnarray}
\end{small}
The training loss of the whole pipeline is the combination of the model space loss $L_{\rm ms}$, image space loss $L_{\rm is}$, and fusion module loss $L_{f}$ which consists of $L_{2}$ loss of $\rm Off$ and $\rm Pos$:
\begin{eqnarray}
L_{f} &=& w_6 \left \| {\rm Off} - {\rm Off}^{\ast} \right \| _{2} + w_7 \left \| {\rm Pos} - {\rm Pos}^{\ast} \right \| _{2}, \\
L &=& L_{\rm ms} + L_{\rm is} + L_{f}.
\end{eqnarray}

\section{Experiments}
\label{sec:exper}

In this section, we evaluate our method's ability to model 3d facial geometry  qualitatively and quantitatively.
We mainly compare our method with methods that have state-of-the-art robustness to occlusion or large view angle, including 3DDFA-V2 \cite{guo2020towards}, SADRNet \cite{ruan2021sadrnet}, SynergyNet \cite{wu2021synergy}.

\subsection{Implement Details}

We train our model end-to-end on the 300W-LP dataset \cite{zhu2016face}, which contains more than 122k synthetic face images across different angles. 
We follow the preprocess and data augmentation of PRN \cite{feng2018prn} and SADRNet \cite{ruan2021sadrnet}, where input images are cropped by their ground truth bounding box and rescaled to size $256\times256$. We select around 45K vertices on the front face region of the model. 
HRNet-W18 \cite{SunXLW19} is adopted as the backbone of the encoder. The whole pipeline has 10.58M trainable parameters, while the image space branch has 5.85M parameters, comparable to the most lightweight 3D dense face alignment algorithms \cite{wu2021synergy, guo2020towards}, but with better performance. 
On average, our method processes one image in 70 ms using a Nvidia RTX 2080 Ti GPU.
During training, the learning rate starts at 5e-6 and reaches 5e-5 after 4 epochs of warm-up. Then, the learning rate decays by a factor of 0.85 every epoch using an exponential scheduler. The training process is optimized by Adam optimizer and lasts 30 epochs where the batch size is set to 12 and weight decay is set to 1e-4. Please refer to the supplementary material for more details.

\subsection{Evaluation Datasets}

We use multiple datasets to evaluate our method's performance on 3D dense face alignment, 3D face reconstruction, and head pose estimation tasks. 

\textbf{AFLW2000-3D} \cite{zhu2016face} is a widely used testing set for 3D dense face alignment, reconstruction and head pose estimation. It contains 2000 unconstrained images. Part of them have occlusion and large poses, so they can be used to evaluate the model's ability to handle challenging cases. We use this dataset to evaluate our model on all three tasks.

\textbf{AFLW2000-3D-occlusion} is a variant of AFLW2000-3D collected by us to especially evaluate a model's robustness to occlusion. It has three subsets: 1. Naturally Occluded Dataset (NOD) contains 127 automatically selected images from AFLW2000-3D. 2. Color Synthetically Occluded Dataset (CSOD) contains 6000 images where every image in AFLW2000-3D is occluded by three different types of color, which are similar to the occlusion patterns in \cite{tiwari2022occlusion}. 3. NatOcc Synthetically Occluded Dataset (NSOD) is generated by Naturalistic Occlusion Generation (NatOcc) technique from \cite{voo2022delving} using full AFLW2000-3D dataset. Details and samples of these subsets are shown in the supplementary material.

\subsection{Face Alignment and Reconstruction}
For this task, we evaluate our method on the AFLW2000-3D benchmark. Following \cite{zhu2016face}, we adopt normalized mean error (NME) where vertices are normalized by the geometric mean of the height and width of the bounding box of the evaluated vertices, as the evaluation metric of face alignment. 
For 2D sparse alignment, we evaluate the NME of 68 facial landmarks on the AFLW2000-3D dataset, shown in \cref{tab:sparse_face_alignment_aflw}, where our method has the best performance.

For 3D dense alignment, we evaluate the NME of the core facial region which has around 45K points on AFLW2000-3D. \cref{tab:dense_alignment} shows that our image space branch solely reaches a new state-of-the-art result. Although the model space branch itself does not have top performance, fusing it with the image space branch is able to achieve a better result. Our whole pipeline outperforms existing methods by a large margin, which is a 5.5\% improvement compared to the existing SOTA method \cite{ruan2021sadrnet}. We notice that the model space feature $f_{ms}$ helps the fusion module to learn how to fuse them together in different cases. Without $f_{ms}$, the fusion module will even hurt the image space branch's performance.

For 3D face reconstruction, we evaluate our method on AFLW2000-3D by NME normalized by outer interocular distance after aligning to the ground truth. 
\cref{tab:dense_alignment} shows our image space branch achieves the best result.

\begin{table}[]
\centering
\resizebox{0.40\textwidth}{!}{
\begin{tabular}{|c|cccc|}
\hline
\multirow{2}{*}{Method} & \multicolumn{4}{c|}{2D Sparse Face Alignment}                                                           \\ \cline{2-5} 
                        & \multicolumn{1}{c|}{0 to 30} & \multicolumn{1}{c|}{30 to 60} & \multicolumn{1}{c|}{60 to 90} & Mean \\ \hline
Dense Corr  \cite{yu2017learning}          & \multicolumn{1}{c|}{3.62}    & \multicolumn{1}{c|}{6.06}     & \multicolumn{1}{c|}{9.56}     & 6.41     \\
3DDFA  \cite{zhu2016face}                 & \multicolumn{1}{c|}{3.78}    & \multicolumn{1}{c|}{4.54}     & \multicolumn{1}{c|}{7.93}     & 5.42     \\
N3DMM    \cite{tran2019learning}      & \multicolumn{1}{c|}{-}       & \multicolumn{1}{c|}{-}        & \multicolumn{1}{c|}{-}        & 4.12     \\
3DSTN       \cite{bhagavatula2017faster}            & \multicolumn{1}{c|}{3.15}    & \multicolumn{1}{c|}{4.33}     & \multicolumn{1}{c|}{5.98}     & 4.49     \\
PRN        \cite{feng2018prn}             & \multicolumn{1}{c|}{2.75}    & \multicolumn{1}{c|}{3.51}     & \multicolumn{1}{c|}{4.61}     & 3.62     \\
DeFA        \cite{liu2017dense}            & \multicolumn{1}{c|}{-}       & \multicolumn{1}{c|}{-}        & \multicolumn{1}{c|}{-}        & 4.50     \\
DAMDN        \cite{jiang2019dual}           & \multicolumn{1}{c|}{2.90}    & \multicolumn{1}{c|}{3.83}     & \multicolumn{1}{c|}{4.95}     & 3.89     \\
SPDT       \cite{piao2019semi}             & \multicolumn{1}{c|}{3.56}    & \multicolumn{1}{c|}{4.06}     & \multicolumn{1}{c|}{\textbf{4.11}}     & 3.88     \\
3DDFA-V2       \cite{guo2020towards}          & \multicolumn{1}{c|}{2.63}    & \multicolumn{1}{c|}{3.42}     & \multicolumn{1}{c|}{4.48}     & 3.51     \\
2DASL        \cite{tu20203d}           & \multicolumn{1}{c|}{2.75}    & \multicolumn{1}{c|}{3.51}     & \multicolumn{1}{c|}{4.61}     & 3.62     \\
SADRNet     \cite{ruan2021sadrnet}            & \multicolumn{1}{c|}{2.66}    & \multicolumn{1}{c|}{3.30}     & \multicolumn{1}{c|}{4.42}     & 3.46     \\
SynergyNet   \cite{wu2021synergy}           & \multicolumn{1}{c|}{2.65}    & \multicolumn{1}{c|}{3.30}     & \multicolumn{1}{c|}{4.27}     & 3.41     \\
\hline
DSFNet-f (ours)            & \multicolumn{1}{c|}{\textbf{2.46}} & \multicolumn{1}{c|}{\textbf{3.20}} & \multicolumn{1}{c|}{4.16}         &   \textbf{3.27}    \\ 
\hline
\end{tabular}}
\caption{Sparse face alignment (68 landmarks) on AFLW2000-3D. The NME (\%) for faces with different yaw angles are reported. }
\label{tab:sparse_face_alignment_aflw}
\end{table}

\begin{table}[]
\centering
\resizebox{0.38\textwidth}{!}{
\begin{tabular}{|c|cc|}
\hline
\multirow{2}{*}{Method} & \multicolumn{2}{c|}{AFLW2000-3D 3D Facial Geometry} \\ \cline{2-3} 
                        & \multicolumn{1}{c|}{Dense Alignment}  & Reconstruction \\ \hline
3DDFA    \cite{zhu2016face}               & \multicolumn{1}{c|}{6.55}         &    5.36  \\
DeFA      \cite{liu2017dense}    & \multicolumn{1}{c|}{6.04}         &    5.64    \\
PRN       \cite{feng2018prn}    & \multicolumn{1}{c|}{4.40}         &    3.96 \\
3DDFA-V2    \cite{guo2020towards} & \multicolumn{1}{c|}{4.18}         &    -  \\
SynergyNet  \cite{wu2021synergy}    & \multicolumn{1}{c|}{4.06}         &    -    \\
SADRNet    \cite{ruan2021sadrnet}   & \multicolumn{1}{c|}{4.02}         &    3.25  \\
\hline
DSFNet-ms (ours)            & \multicolumn{1}{c|}{4.15}         &     3.46     \\ 
DSFNet-is (ours)            & \multicolumn{1}{c|}{3.89}         &      \textbf{3.16}    \\ 
DSFNet-f (ours)   w/o $f_{ms}$         & \multicolumn{1}{c|}{3.96}         &     3.27     \\ 
DSFNet-f (ours)            & \multicolumn{1}{c|}{\textbf{3.80}}         &     3.24     \\ 
\hline
\end{tabular}}
\caption{3D dense face alignment and reconstruction results on AFLW2000-3D. NME (\%) normalized by bounding box side length and NME (\%) normalized by 3D outer interocular distance are reported separately for 3D dense alignment and 3D reconstruction tasks. $-ms$, $-is$, $-f$ denote our method's model space prediction, image space prediction, and fusion model prediction separately, where all modules are trained jointly in the fusion model.}
\label{tab:dense_alignment}
\end{table}

\subsection{Head Pose Estimation}
For head pose estimation task, following the metric in \cite{wu2021synergy, cao2021vector}, we compute the mean absolute error (MAE) of each Euler angle and the MAE of all three Euler angles. We evaluate our method on images with yaw angle  ranged from -99 degrees to 99 degrees in AFLW2000-3D. \cref{tab:head_pose} shows the superiority of our method. In particular, our method infers much more accurate yaw angle prediction compared to existing methods.

\begin{table}[]
\centering
\resizebox{0.35\textwidth}{!}{
\begin{tabular}{|c|c|c|c|c|}
\hline
AFLW2000-3D  & Yaw  & Pitch & Roll & Mean \\ \hline
FSANet  \cite{yang2019fsa}     & 4.50 & 6.08  & 4.64 & 5.07 \\
TriNet   \cite{cao2021vector}    & 4.20 & 5.77  & 4.04 & 3.97 \\
RankPose  \cite{dai2020rankpose}   & 2.99 & 4.75  & 3.25 & 3.66 \\
3DDFA    \cite{zhu2016face}    & 4.33 & 5.98  & 4.30 & 4.87 \\
2DASL     \cite{tu20203d}   & 3.85 & 5.06  & 3.50 & 4.13 \\
FDN     \cite{zhang2020fdn}     & 3.78 & 5.61  & 3.88 & 4.42 \\
MNN     \cite{valle2020multi}     & 3.34 & 4.69  & 3.48 & 3.83 \\
3DDFA-V2  \cite{guo2020towards}   & 4.06 & 5.26  & 3.48 & 4.27 \\
WHENet-V  \cite{zhou2020whenet}   & 4.44 & 5.75  & 4.31 & 4.83 \\
img2pose  \cite{albiero2021img2pose}   & 3.43 & 5.03  & 3.28 & 3.91 \\
SADRNet  \cite{ruan2021sadrnet}    & 2.93 & 5.00  & 3.54 & 3.82 \\
SynergyNet  \cite{wu2021synergy} & 3.42 & \textbf{4.09} & \textbf{2.55} & 3.35 \\
6DRepNet   \cite{hempel20226d}  & 3.63 & 4.91  & 3.37 & 3.97 \\
\hline
DSFNet-f (ours) & \textbf{2.65} & 4.28  & 2.82  & \textbf{3.25} \\ 
\hline
\end{tabular}}
\caption{Head pose estimation results on AFLW2000-3D dataset.}
\label{tab:head_pose}
\end{table}

\subsection{Occlusion Robustness}
Benefiting from the image space branch, our method is robust to occlusion and large view angles as shown in \cref{fig:vis}. Quantitative evaluation on the three occlusion-specific subsets (i.e., one natural occlusion, NOD, and two synthetic occlusion subsets, CSOD and NSOD) is performed to further demonstrate the superior robustness of the proposed method compared with the SOTA methods as shown \cref{tab:occ}. 

\begin{table}[]
\centering
\resizebox{0.32\textwidth}{!}{
\begin{tabular}{|c|ccc|}
\hline
Method & NOD  & CSOD & NSOD \\ \hline
SynergyNet \cite{wu2021synergy}   & 5.49 &  7.95     &   14.66  \\
SADRNet \cite{ruan2021sadrnet}   & 6.70 &  5.31     &   8.70  \\
DSFNet-ms (ours)   & 5.23 &  5.06     &   6.62  \\ 
DSFNet-is (ours)   & \textbf{4.78} & 4.59      &   5.83  \\
DSFNet-f (ours)   & 4.80 &  \textbf{4.44}     &   \textbf{5.40}  \\ \hline
\end{tabular}}
\caption{Occlusion resistance results on AFLW2000-3D-occlusion. NME (\%) of 3d dense face alignment is reported.}
\label{tab:occ}
\end{table}

\subsection{Complementary Relationship between Image Space and Model Space Branches}
\label{sec:dual_space}

Although the image space branch solely already achieves the SOTA accuracy and occlusion robustness, it has some drawbacks. In our framework, the model space branch complements the image space branch as they focus on different information. The image space can handle large pitch and roll angles where existing methods fail, while it has more difficulty handling large yaw angles. The reason is straightforward: In large yaw angle cases, most of the visible region is in the side face where fewer facial attributes lay, which leads the image space branch hard to deduce the accurate UV locations of visible pixels. Second, we train our model on 300W-LP which is a synthetic dataset consisting of images generated by 3D rotation around the yaw angle from the positive images, resulting in artifacts in the side face region. Since the image space branch relies more on low-level image information, it is sensitive to this data problem. In contrast, the model space branch focuses more on high-level information, and is more robust to large yaw angles. Similarly, due to sensitivity to low-level information, the image space is easier to be interfered by noise and blur. In this situation, model space prediction is an important complementary for the robustness of the whole pipeline.

\subsection{Ablation Study of the Post-process Algorithm}
\label{sec:abla_post}
To evaluate the contribution of each component in the proposed post-process algorithm, an ablation study is performed using the image space branch on AFLW2000-3D dataset as shown in \cref{tab:post_process}. It is observed that disentangling the prediction of $T_{is}$ and $\alpha_{is}$ by the $\rm Align$ algorithm and the PointNet separately is crucial to the high performance of the post-process algorithm.

\begin{figure}[t]
  \centering
   \includegraphics[width=1\linewidth]{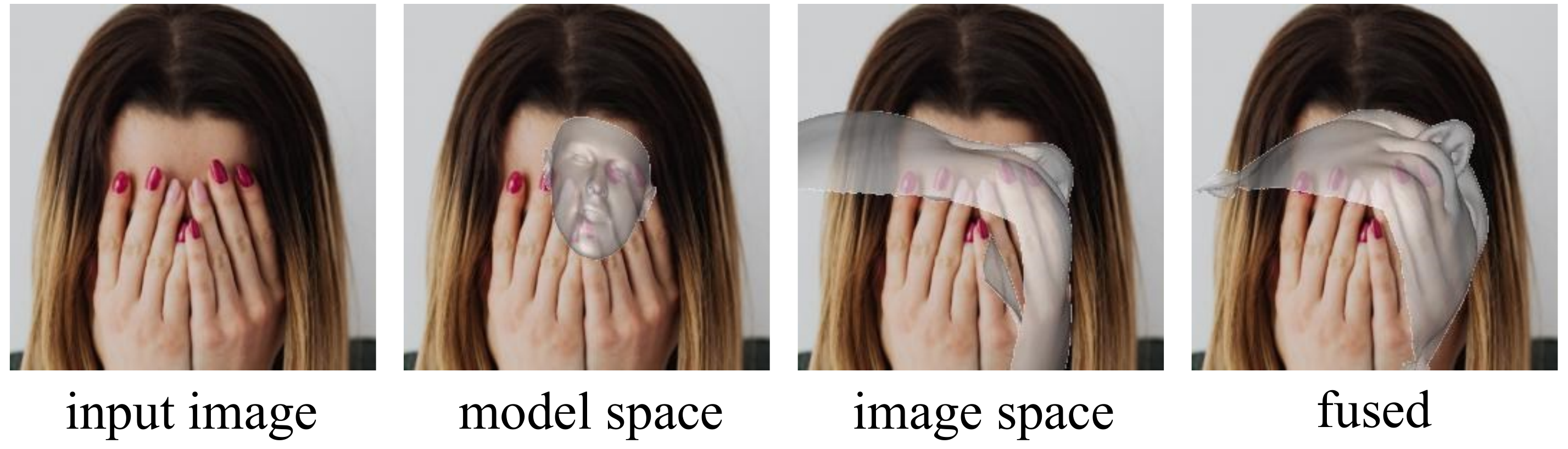}
   \vspace{-5mm}
   \caption{Failure case of our method.}
   \label{fig:rebuttal_fig2}
\end{figure}

\subsection{Failure Cases and Limitations}
\cref{fig:rebuttal_fig2} shows a typical failure case of our method. The pose and scale of our image space prediction rely on the number of the points sampled from the visible facial region. If the visible facial region is too small, the valid points will be too few. Thus, while our model space predicts the coarse pose based on the context information like the hair, image space prediction fails due to the small visible facial area, resulting in the wrong final prediction. 

Besides, our method adopts a highly supervised training framework and relies on the quality of the labeled data. To address the limitation, we intend to adopt semi-supervised or self-supervised learning to make use of the unlabeled data in future work.  
Another issue is the assumption of orthographic projection, where most methods trained on 300W-LP \cite{zhu2016face}, including ours, use weak perspective or orthographic projection. This limits our method in handling images taken from a close distance. Exploring the use of perspective projection \cite{kao2022single} is an area for future work.

\begin{table}[]
\centering
\resizebox{0.34\textwidth}{!}{
\begin{tabular}{|c|c|c|c|c|c|c|}
\hline
\textit{xyz} & $Cor$ & $Cf$ & $L_{cons}$ & $Align$ & NME & $diff$ \\ \hline
 $\checkmark$   &    &   & $\checkmark$    &     $\checkmark$   &    4.01 & 0.12 \\
    $\checkmark$ &   $\checkmark$  &   &   $\checkmark$  &    $\checkmark$  &   3.93 & 0.04  \\
   $\checkmark$  &   $\checkmark$  &  $\checkmark$  &     &     $\checkmark$   &   3.94  & 0.05 \\
    $\checkmark$ &  $\checkmark$   &   $\checkmark$ &   $\checkmark$   &       &   4.12  & \textbf{0.23} \\
   $\checkmark$  &   $\checkmark$  &   $\checkmark$ &     $\checkmark$ &      $\checkmark$  &  \textbf{3.89}   & - \\ \hline
\end{tabular}}
\caption{3D dense face alignment results of our image space branch on AFLW2000-3D dataset with different post-process modules. The first three are the ablation study of the input components of the PointNet. \textit{xyz},  $Cor$, $Cf$ denotes using the point cloud $Pc$'s 3D locations from depth map, UV locations from correspondence map, and values from confidence map as the PointNet's input. $L_{cons}$ denotes using $L_{cons}$ in training. $Align$ denotes if the point cloud is aligned to the canonical view or not before being fed to the PointNet. }
\label{tab:post_process}
\end{table}

\vspace{0.16cm}
\section{Conclusions}
\label{sec:concl}
We have presented a dual space fusion network (DSFNet) for robust 3D facial geometry modeling in unconstrained scenarios. Instead of directly regressing the whole face, our image space branch predicts a 3D facial geometry's 2D image space representation, which decentralizes the instance-level prediction to pixel-level dense prediction. A novel PointNet-based post-process algorithm is proposed to recover the face's whole 3D geometry from the visible region information contained in the image space representation. Since it makes better use of low-level 2D spatial and semantic information, it achieves high robustness to occlusion and large view angles.
We further proposed a fusion module to combine advantages from both image and model space branches, resulting in a new state-of-the-art performance in mainstream benchmarks on 3D dense face alignment and reconstruction, and head pose estimation tasks. 

\vspace{0.5cm}
\noindent {\bf Acknowledgments}
\label{sec:ack}
This research is supported by the National Research Foundation, Singapore under its Strategic Capability Research Centres Funding Initiative. Any opinions, findings and conclusions or recommendations expressed in this material are those of the author(s) and do not reflect the views of National Research Foundation, Singapore.

\clearpage

{\small
\bibliographystyle{ieee_fullname}
\bibliography{egbib}
}

\end{document}